\documentclass[10pt]{article} % For LaTeX2e

\usepackage[preprint]{rlj} % Should be uncommented for preprint versions

\usepackage{amssymb}            % Defines common symbols like \mathbb R
\usepackage{mathtools}          % Extends amsmath, providing common math tools
\usepackage{mathrsfs}           % Enables \mathscr, which can work in cases that \mathcal does not
\usepackage{graphicx}           % For including images
\usepackage{subcaption}         % Allows for the use of subfigures and subcaptions
\usepackage[space]{grffile}     % For spaces in image names
\usepackage{url}                % For displaying URLs
\usepackage{lipsum}             % For placeholder text

\usepackage{tabularx}
\usepackage{booktabs}
\usepackage{multirow}
\usepackage{threeparttable}

\usepackage{wrapfig} % 允许表格浮动
\usepackage{lipsum}  % 生成示例文本，可移除
\title{Residual Policy Gradient: \\ A Reward View of KL-regularized Objective}

\setrunningtitle{Residual Policy Gradient}

\author{Pengcheng Wang, % \textsuperscript, 
Xinghao Zhu, 
Yuxin Chen,
Chenfeng Xu,
Masayoshi Tomizuka,
Chenran Li
}
\emails{wangpc@berkeley.edu, chenran\_li@berkeley.edu}
\affiliations{
\textbf{Department of Mechanical Engineering, University of California, Berkeley}\\
}

\contribution{
    We derive a concise form of maximum-entropy policy gradient method, \emph{Soft Policy Gradient}, as a preliminary and compare it with other forms in existing literature,
    }
    {
    Some previous works have also focused on deriving the max entropy policy gradient. \cite{williams1992simple} and \cite{  mnih2016asynchronous} only considers the first-step impact of the entropy; \cite{schulman2017equivalence} brings the entropy to every step but repeats the first-step entropy; \cite{shi2019soft} derives the same result but does not formalize it to the simplest form.
    }

\contribution{
    Building on this, we introduce \emph{Residual Policy Gradient} (RPG), which extends RQL to policy gradient methods, allowing policy customization in gradient-based RL settings.
    }
    {
    RQL~\citep{RQL} proposes a principled framework to solve policy customization without the reward knowledge of the prior policy. 
    This work extends the RQL to policy gradient methods for broader application.
    }

\contribution{
    With the view of RPG, we rethink the KL-regularized objective widely used in RL fine-tuning, showing that (1) If the prior policy models an optimal Boltzmann distribution, the KL-regularized objective induces a maximum-entropy policy that balances inherent properties and task-specific requirements at the reward level.
    (2) This formulation can be further refined through a simple decoupling, leading to improved optimization performance.
    }
    {
    The KL-regularized objective is related to KL Control~\citep{todorov2006linearly, kappen2012optimal}, which aims to prevent the target policy from deviating too far from a reference, which is widely adopted in RL~\citep{rawlik2013stochastic, fox2015taming}. The flexibility of KL decoupling is also reflected in many works~\citep{teh2017distral, lin2025td}.
    }

 % Your keywords

\summary{% Policy Customization is important
Reinforcement Learning and Imitation Learning have achieved widespread success in many domains but remain constrained during real-world deployment. One of the main issues is the additional requirements that were not considered during training.
To address this challenge, policy customization has been introduced, aiming to adapt a prior policy while preserving its inherent properties and meeting new task-specific requirements. 
A principled approach to policy customization is Residual Q-Learning (RQL), which formulates the problem as a Markov Decision Process (MDP) and derives a family of value-based learning algorithms. 
However, RQL has not yet been applied to policy gradient methods, which restricts its applicability, especially in tasks where policy gradient has already proven more effective. 
In this work, we first derive a concise form of \emph{Soft Policy Gradient} as a preliminary. 
Building on this, we introduce \emph{Residual Policy Gradient} (RPG), which extends RQL to policy gradient methods, allowing policy customization in gradient-based RL settings.
With the view of RPG, we rethink the KL-regularized objective widely used in RL fine-tuning. We show that under certain assumptions, KL-regularized objective leads to a maximum-entropy policy that balances the inherent properties and task-specific requirements on a reward-level.
Our experiments in MuJoCo demonstrate the effectiveness of \emph{Soft Policy Gradient} and \emph{Residual Policy Gradient}.

}

\begin{document}

% \makeCover  % Create the cover page
\maketitle  % Make the title section

\begin{abstract}
% Policy Customization is important
Reinforcement Learning and Imitation Learning have achieved widespread success in many domains but remain constrained during real-world deployment. One of the main issues is the additional requirements that were not considered during training.
% Definition of Policy Customization
To address this challenge, policy customization has been introduced, aiming to adapt a prior policy while preserving its inherent properties and meeting new task-specific requirements. 
% RQL can solve that
A principled approach to policy customization is Residual Q-Learning (RQL), which formulates the problem as a Markov Decision Process (MDP) and derives a family of value-based learning algorithms. 
% it is not enough
However, RQL has not yet been applied to policy gradient methods, which restricts its applicability, especially in tasks where policy gradient has already proven more effective. 
% Contribution 1:
In this work, we first derive a concise form of \emph{Soft Policy Gradient} as a preliminary. 
% Contribution 2:
Building on this, we introduce \emph{Residual Policy Gradient} (RPG), which extends RQL to policy gradient methods, allowing policy customization in gradient-based RL settings.
% Contribution 3:
With the view of RPG, we rethink the KL-regularized objective widely used in RL fine-tuning. We show that under certain assumptions, KL-regularized objective leads to a maximum-entropy policy that balances the inherent properties and task-specific requirements on a reward-level.
% Experiments
Our experiments in MuJoCo demonstrate the effectiveness of \emph{Soft Policy Gradient} and \emph{Residual Policy Gradient}.

\end{abstract}

\section{Introduction}\label{sec:intro}

%%% Policy Customization is Important
% General Policy Learning
Policy learning algorithms, such as Reinforcement Learning (RL) and Imitation Learning (IL), have achieved great success across various cross-domain tasks, including locomotion~\citep{liang2024adaptive}, autonomous racing~\citep{Sophy}, and Large Language Model (LLM)~\citep{guo2025deepseek}. 
% Additional Requirements
However, practical deployments often impose additional requirements on the trained policies beyond those established during training, such as configuring the robot-human hand-over~\citep{cakmak2011human} or membership behaviour~\citep{correia2019choose} with preferences. 
% % Policy Customization is essential to meet this need
% Under this context, retraining a policy network from scratch for each new objective can be both costly and inefficient. 
% Policy Customization introduction
To elaborate the challenge, \textit{policy customization}, introduced by \citet{RQL}, aims to modify a prior policy, ensuring that the new policy (1) retains properties of the original and (2) adapts to new task-specific requirements.

% RQL Introduction and Policy Gradients Limitation 1
Residual Q-Learning (RQL)~\citep{RQL} provides a principled approach to policy customization by formulating the problem as a Markov Decision Process (MDP), which jointly optimizies the reward of the prior and the add-on task-sepecift reward. It derivies a theoretical learning objective through the Q-function and provides a family of pratical learning and planning algorithms.
% However, it has primarily been applied in value-based RL and has not yet been extended to policy gradient methods, a fundamental class of algorithms in modern reinforcement learning.
% % Example
% This limitation restricts the applicability of RQL, especially when large-scale parallel environments are available, policy gradient methods tend to outperform value-based methods~\citep{gallici2024simplifying}.
However, it has primarily been applied in value-based RL and has not yet been extended to policy gradient methods, a fundamental class of algorithms in modern reinforcement learning, which restricts the applicability of RQL.
% Example
This limitation becomes more pronounced when large-scale parallel environments are available, where policy gradient methods tend to outperform value-based methods~\citep{gallici2024simplifying}.
% So it is important
Therefore, it is of great importance to derive the policy gradient extension of RQL.
% Contribution1&2: Soft PPO & Residual PPO

% Soft Policy Gradient
In this work, we first derive a concise form of maximum-entropy policy gradient method, \emph{Soft Policy Gradient}, as the preliminary method and compare it with similar forms in the related literature.
% Residual Policy Gradient
Based on the derived form, we further propose \emph{Residual Policy Gradient} (RPG) to solve the policy customization in policy gradient approach. Our experiments in MuJoCo demonstrate the effectiveness of \emph{Soft Policy Gradient} and \emph{Residual Policy Gradient}.
% KL
Moreover, with the view of RPG, we rethink the widely used KL-regularized objective in RL fine-tuning. We show that:
(1) If the prior policy models an optimal Bolzmann distribution, the KL-regularized objective induces a maximum-entropy policy that balances inherent properties and task-specific requirements at the reward level.
(2) This formulation can be further refined through a simple decoupling, leading to improved optimization efficiency.
% RLHF example
In Appendix~\ref{appendix: RLHF}, we elaborate these insights with RL-based LLM fine-tuning, demonstrating their implications for both theoretical understanding and future research directions.

% % Bring in the RLHF
% As a canonical and essential example of policy customization, Reinforcement Learning from Human Feedback (RLHF) also aims to fine-tune models based on human preferences while preserving core capabilities.
% % RLHF Situation
% Despite the great empirical success achieved by many existing RLHF approaches, they often stem from simple intuitions~\citep{ziegler2019fine} or rely on similar update mechanisms without a unified theoretical foundation~\citep{rafailov2024direct, rafailov2024r, li2024q, yuan2024implicitprm}.
% % Contribution
% In this work, we show that RPG serves as a unified framework, revealing that many existing RLHF methods can be interpreted as optimizing a maximum-entropy policy on the same token-level MDP.
% % Future
% The success of the mentioned RLHF works also serves as a solid empirical validation for our theoretical framework, which lays a foundation for future RLHF and broader policy customization research.

\section{Preliminaries}\label{sec:Preliminaries}
In this section, we introduce two techniques mainly used in the proposed algorithms, Policy Gradient and RQL, laying the foundation for the subsequent derivations and analysis.

\subsection{Policy Gradient Method}\label{sec:PG}

Policy gradient method is one of the biggest families of the RL algorithm, which directly optimizes the policy $\pi_\theta$ on the expected return $J$ of a discounted reward function $r$ by estimating its gradient:

\begin{equation}\label{eqn:PG obj}
\nabla_\theta J(\pi_\theta) = \nabla_\theta \mathbb{E}_{(\boldsymbol{s}_t,\boldsymbol{a}_t)\sim\pi_\theta,p(\cdot|\boldsymbol{s},\boldsymbol{a})}\left[\sum_{t=0} \gamma^t \left(r(\boldsymbol{s}_t,\boldsymbol{a}_t)\right)\right],
\end{equation}

where $\gamma$ is the discounted factor, and $p(\cdot|\boldsymbol{s},\boldsymbol{a})$ is the transition probability of the environment. Based on Eqn~\ref{eqn:PG obj}, various practical algorithms have been developed to effectively address a wide range of tasks. A typical example is PPO~\citep{schulman2017proximal}, which ensures stable and reliable updates by introducing a clipped objective function:

\begin{equation}\label{eqn:PPO obj}
\quad \nabla_\theta J(\pi_\theta) = \nabla_\theta \mathbb{E}_{(\boldsymbol{s}_t,\boldsymbol{a}_t)\sim\pi_\theta,p(\cdot|\boldsymbol{s},\boldsymbol{a})} \left[ \min \{r_t(\pi_\theta) A_t, \operatorname{clip}(r_t(\pi_\theta), 1 - \epsilon, 1 + \epsilon) A_t \} \right],
\end{equation}
where $r_t(\pi_\theta) = \frac{\pi_\theta(\boldsymbol{a}_t | \boldsymbol{s}_t)}{\pi_{\theta_{old}}(\boldsymbol{a}_t | \boldsymbol{s}_t)}$ denotes the probability ratio, $A_t$ is the estimator of the advantage function at timestep $t$, and $\epsilon$ is a hyperparameter. This augmented form limits policy updates within a pre-defined range, balancing exploration and exploitation without complex constraints.

\subsection{Policy Customization and Residual Q-Learning}\label{sec:RQL}
% The goal of policy customization is to derive a policy from a prior policy that satisfies the new requirements of an add-on task while maintaining the performance of the original task. Thus, it is assumed that a well-trained prior policy is available for policy customization. 
Consider a MDP defined by $\mathcal{M} = (\mathcal{S}, \mathcal{A}, r, p)$, where $\mathcal{S}$ and $\mathcal{A}$ are state and action spaces, $r$ is the reward function, and $p$ is the transition probability density function. The objective of policy customization is to derive a new policy that optimally solves a modified MDP, $\hat{\mathcal{M}} = (\mathcal{S}, \mathcal{A}, \omega r+r_R, p)$, where $r_R$ is the add-on task-specified reward and $\omega$ is the weight parameter. In particular, policy customization is especially relevant in scenarios where the basic reward $r$ is unknown, but a prior policy $\pi$, trained as an optimal maximum-entropy policy for $\mathcal{M}$, is available.

%Meanwhile, the add-on task is specified by an add-on reward function $r_R$ and the full task is a new MDP defined by $\hat{\mathcal{M}} = (\mathcal{S}, \mathcal{A}, \omega r+r_R, p)$, where the reward is a weighted sum of the basic reward $r$ and the add-on reward $r_R$. The goal of policy customization is to find a policy that solves this new MDP. 
% While the formulation applies whether the reward function $r$ is given or not, we are primarily interested in the case when $r$ remains \emph{unknown} to ensure broader applicability across different prior policies obtained through various methods, including those solely imitating demonstrations.

\cite{RQL} proposed the residual Q-learning (RQL) framework to solve the policy customization. 
Given the prior policy $\pi$, RQL is able to find this customized policy without the reward knowledge of the prior policy $r$, which ensures broader applicability across different prior policies obtained through various methods, including those solely imitating demonstrations. Specifically, by defining the Residual Q function $Q_R = \hat{Q} - \omega Q$, where $\hat{Q}$ and $Q$ is the optimal value function on $\hat{\mathcal{M}}$ and $\mathcal{M}$, RQL updates the $Q_R$ using the following equation:
\begin{equation}
\label{eqn: RQL}
    Q_R(\boldsymbol{s}, \boldsymbol{a}) \leftarrow r_R(\boldsymbol{s}, \boldsymbol{a}) + \gamma \mathbb{E}_{s'}\left[\hat{\alpha}\log \int_{\mathcal{A}}\exp\left(\frac{1}{\hat{\alpha}}\left({Q}_{R} (\boldsymbol{s}', \boldsymbol{a}')+\omega\alpha\log \pi\left(\boldsymbol{a}' \vert \boldsymbol{s}'\right)\right)\right)d\boldsymbol{a}'\right],
\end{equation}
where $\alpha$ and $\hat{\alpha}$ are the entropy-encouragement coefficients of maximum-entropy policy on prior and full task, and $\gamma$ is the discounted factor.
As shown in the RQL~\citep{RQL} appendix, by adding $\omega \alpha \log \pi\left(\boldsymbol{a}_{t}| \boldsymbol{s}_{t}\right)$ to both side of Eqn.~\ref{eqn: RQL} and defining $Q^{aug}\left(\boldsymbol{s}, \boldsymbol{a}\right) = Q_R\left(\boldsymbol{s}, \boldsymbol{a}\right) + \omega\alpha\log \pi\left(\boldsymbol{a} \vert \boldsymbol{s}\right)$, we can reformulate Eqn.~\ref{eqn: RQL} into:
\begin{equation}
\label{eqn: RQL_aug}
    Q^{aug}(\boldsymbol{s}, \boldsymbol{a}) \leftarrow r_R(\boldsymbol{s}, \boldsymbol{a}) + \omega\alpha\log \pi\left(\boldsymbol{a} \vert \boldsymbol{s}\right) + \gamma \mathbb{E}_{s'}\left[\hat{\alpha}\log \int_{\mathcal{A}}\exp\left(\frac{1}{\hat{\alpha}}\left({Q}^{aug} (\boldsymbol{s}', \boldsymbol{a}')\right)\right)d\boldsymbol{a}'\right],
\end{equation}

which is a standard soft-Q update on an augmented MDP problem $\mathcal{M}^\mathrm{aug} = (\mathcal{S}, \mathcal{A}, \omega^\prime \log \pi + r_R, p)$. This indicates that we can find the maximum-entropy policy solving the full MDP problem $\hat{\mathcal{M}}$ by solving an augmented MDP problem $\mathcal{M}^\mathrm{aug}$, where $\omega^\prime = \omega \alpha$ is a parameter that balances the optimality between basic and add-on tasks.

\section{Residual Policy Gradient}\label{sec:residual_PG}
In this section, we first derive a more netural form of the maximum-entropy policy gradient methods \textit{Soft Policy Gradients}, which leads to \textit{Residual Policy Gradient} (RPG). Furthermore, we propose \textit{Soft PPO} and \textit{Residual PPO} as two practical algorithms for deployment.

\subsection{Soft Policy Gradient}\label{sec:SPG}
% Motivation
As discussed in Sec.~\ref{sec:RQL}, to utilize the augmented MDP problem $\mathcal{M}^\mathrm{aug}$, we need to extend the policy gradient methods to optimize the maximum-entropy policy.

Let $\tau = (\boldsymbol{s}_0, \boldsymbol{a}_0, \boldsymbol{s}_1, \boldsymbol{a}_1, \dots, \boldsymbol{s}_t, \boldsymbol{a}_t, \dots)$ represents a complete trajectory, and $p_\theta(\tau)$ represents the probability density of $\tau$ generated by a paramterized policy $\pi_\theta$ and the environment dynamics $p(\cdot|\boldsymbol{s},\boldsymbol{a})$. The maximum-entropy objective can be written as:
\begin{equation}\label{eqn:MERL_obj}
J(\pi_\theta) = \mathbb{E}_{\tau \sim p_\theta(\tau)}\left[\sum_{t=0} \gamma^t \left(r(\boldsymbol{s}_t,\boldsymbol{a}_t)+\alpha \mathbb{E}_{\boldsymbol{a}_t\sim\pi_\theta(\cdot|\boldsymbol{s}_t)}\left[-\log\pi_\theta\left(\boldsymbol{a}_t|\boldsymbol{s}_t\right)\right]\right)\right].
\end{equation}

The estimation of its gradient $\nabla_\theta J(\pi_\theta)$ can be derived as:
\begin{equation}
\label{eqn: SPG}
\mathbb{E}_{\tau \sim p_\theta(\tau)} \left[ \sum_{t=0} \nabla_\theta \log \pi_\theta\left(\boldsymbol{a}_t|\boldsymbol{s}_t\right) \left(\sum_{t^\prime = t} \gamma^{t^\prime - t} \left(
    r(\boldsymbol{s}_{t^\prime},\boldsymbol{a}_{t^\prime})-\alpha\log\pi_{\theta}\left(\boldsymbol{a}_{t^\prime}|\boldsymbol{s}_{t^\prime}\right)\right) \right) \right]
\end{equation}

% \begin{equation}
% \mathbb{E}_{\tau \sim p_\theta(\tau)} \left[ \sum_{t=0}
% \nabla_\theta \log \pi_\theta (\boldsymbol{a}_t | \boldsymbol{s}_t) \left( \sum_{t^\prime=t} \gamma^{t^\prime-t} \left(r(\boldsymbol{s}_{t^\prime},\boldsymbol{a}_{t^\prime}) - \alpha\log \pi_\theta(\boldsymbol{a}_{t^\prime} | \boldsymbol{s}_{t^\prime})\right) - V^{\pi_\theta}(\boldsymbol{s}_t) \right)  \right],
% \end{equation}

% where $V^{\pi_\theta}(\boldsymbol{s}_{t}) = \mathbb{E}_{\tau \sim p_\theta(\tau)} \left[\sum_{t^\prime=t} \gamma^{t^\prime-t} \left(r(\boldsymbol{s}_{t^\prime},\boldsymbol{a}_{t^\prime})-\alpha\log\pi_{\theta}\left(\boldsymbol{a}_{t^\prime}|\boldsymbol{s}_{t^\prime}\right)\right)\right]$ is the value function. 
This is equivalent to applying policy gradient over a reformulated reward $r(\boldsymbol{s}_{t},\boldsymbol{a}_{t})-\alpha\log\pi_{\theta}\left(\boldsymbol{a}_{t}|\boldsymbol{s}_{t}\right)$.
A detailed derivation can be found in Appendix~\ref{appendix: derivation}. 
Although some previous works \citep{williams1992simple, mnih2016asynchronous, shi2019soft, schulman2017equivalence} also focused on deriving the maximum-entropy policy gradient, their formulations come with certain inconsistencies or implicit assumptions. With a unified notation, we compare and highlight their similarities and differences.

In the naive entropy regularized policy gradient methods \citep{williams1992simple, mnih2016asynchronous}, the gradient estimator is encouraged with only an entropy term at the end, which is:
\begin{equation}
\label{eqn: end-entropy}
\mathbb{E}_{\tau \sim p_\theta(\tau)} \left[ \sum_{t=0}
\nabla_\theta \log \pi_\theta (\boldsymbol{a}_t | \boldsymbol{s}_t) \left( \sum_{t^\prime=t} \gamma^{t^\prime-t} r(\boldsymbol{s}_{t^\prime},\boldsymbol{a}_{t^\prime})
% - V^{\pi_\theta}(\boldsymbol{s}_t) 
\right) - \alpha \nabla_\theta \mathcal{H}\left(\pi_\theta \left(\cdot | \boldsymbol{s}_t\right) \right)  \right].
\end{equation}

Note that $\mathbb{E}_{\tau \sim p_\theta(\tau)} \left[ \sum_{t=0}
\nabla_\theta \mathcal{H}\left(\pi_\theta \left(\cdot | \boldsymbol{s}_t\right) \right) \right] = \mathbb{E}_{\tau \sim p_\theta(\tau)} \left[ \sum_{t} \nabla_\theta \log \pi_\theta(\boldsymbol{a}_t | \boldsymbol{s}_t) \left(\log \pi_\theta(\boldsymbol{a}_t | \boldsymbol{s}_t) \right) \right]$, we can reformulate the naive gradient estimator with a delta function:
\begin{equation}
\begin{aligned}
\mathbb{E}_{\tau \sim p_\theta(\tau)} \left[ \sum_{t=0} 
\nabla_\theta \log \pi_\theta (\boldsymbol{a}_t | \boldsymbol{s}_t) \left( \sum_{t^\prime=t} \gamma^{t^\prime-t} \left(r(\boldsymbol{s}_{t^\prime},\boldsymbol{a}_{t^\prime}) - \alpha \delta(t^\prime, t)\log \pi_\theta(\boldsymbol{a}_{t^\prime} | \boldsymbol{s}_{t^\prime}) \right) 
% - V^{\pi_\theta}(\boldsymbol{s}_t) 
\right)  \right],
\end{aligned}
\end{equation} 
which indicates that the naive methods only consider the first-step ($t^\prime = t$) entropy bonus in updating.

In \cite[Eqn.(80)]{schulman2017equivalence}, the author introduced the proper entropy bonus on every step to the expected return but retained the $- \alpha \nabla_\theta \mathcal{H}\left(\pi_\theta \left(\cdot | \boldsymbol{s}_t\right) \right)$ term, leading to an update as:
\begin{equation}
\label{eqn: repeat-entropy}
\mathbb{E}_{\tau \sim p_\theta(\tau)} \left[ \sum_{t=0}
\nabla_\theta \log \pi_\theta (\boldsymbol{a}_t | \boldsymbol{s}_t) \left( \sum_{t^\prime=t} \gamma^{t^\prime-t} \left( r(\boldsymbol{s}_{t^\prime},\boldsymbol{a}_{t^\prime}) - \alpha (\delta(t^\prime, t)+1)\log \pi_\theta(\boldsymbol{a}_{t^\prime} | \boldsymbol{s}_{t^\prime})\right) 
% - V^{\pi_\theta}(\boldsymbol{s}_t) 
\right)  \right],
\end{equation}
which repeats the entropy bonus on the first step.

In \cite[Eqn.(13)]{shi2019soft}, the derived gradient estimator is
\begin{equation}
\label{eqn: tsinghua derivation}
    \mathbb{E}_{\tau \sim p_\theta(\tau)} \left[\sum_{t=0} \nabla_\theta \log \pi_\theta\left(\boldsymbol{a}_t|\boldsymbol{s}_t\right) \left(Q^{\pi_\theta}(\boldsymbol{s}_{t},\boldsymbol{a}_{t}) - \alpha\log\pi_{\theta}\left(\boldsymbol{a}_{t}|\boldsymbol{s}_{t}\right) - 1 \right)   \right].
\end{equation}
Since     $\mathbb{E}_{\tau \sim p_\theta(\tau)} \left[\sum_{t=0} \nabla_\theta \log \pi_\theta\left(\boldsymbol{a}_t|\boldsymbol{s}_t\right)   \right] = 0$, we can eliminate the $1$ in Eqn.~\ref{eqn: tsinghua derivation} which leads to our result by substituting $Q^{\pi_\theta}(\boldsymbol{s}_{t},\boldsymbol{a}_{t}) = r(\boldsymbol{s}_{t},\boldsymbol{a}_{t}) + \sum_{t^\prime > t} \gamma^{t^\prime-t} \left(
    r(\boldsymbol{s}_{t^\prime},\boldsymbol{a}_{t^\prime})-\alpha\log\pi_{\theta}\left(\boldsymbol{a}_{t^\prime}|\boldsymbol{s}_{t^\prime}\right)\right)$.

\subsection{Practical Algorithm}
\label{sec: practical algorithm}
\textbf{Soft PPO.}
By applying advantage function, we can reform the gradient:
\begin{equation}\label{eqn:Final_SPG}
    \nabla_\theta J(\pi_\theta) = \mathbb{E}_{\tau \sim p_\theta(\tau)} \left[ \sum_{t=0} \nabla_\theta \log \pi_\theta\left(\boldsymbol{a}_t|\boldsymbol{s}_t\right) A^{\mathrm{SPG}}_{\pi_\theta}(\boldsymbol{s}_t,\boldsymbol{a}_t) \right],
\end{equation}
where
\begin{equation}
    A^{\mathrm{SPG}}_{\pi_\theta}(\boldsymbol{s}_t,\boldsymbol{a}_t) = r(\boldsymbol{s}_{t},\boldsymbol{a}_{t})-\alpha\log\pi_{\theta}\left(\boldsymbol{a}_{t}|\boldsymbol{s}_{t}\right) + \gamma \mathbb{E}_{\boldsymbol{s}_{t+1}\sim p(\cdot|\boldsymbol{s}_t,\boldsymbol{a}_t)}  [V^{\pi_\theta}(\boldsymbol{s}_{t+1})] - V^{\pi_\theta}(\boldsymbol{s}_{t}),
\end{equation}
\begin{equation}
    V^{\pi_\theta}(\boldsymbol{s}_{t}) = \mathbb{E}_{\tau \sim p_\theta(\tau)} \left[\sum_{t^\prime=t} \gamma^{t^\prime-t} \left(
    r(\boldsymbol{s}_{t^\prime},\boldsymbol{a}_{t^\prime})-\alpha\log\pi_{\theta}\left(\boldsymbol{a}_{t^\prime}|\boldsymbol{s}_{t^\prime}\right)\right)\right],
\end{equation}
which is equivalent to applying classic policy gradient over $r^{\mathrm{SPG}} = r(\boldsymbol{s}_{t},\boldsymbol{a}_{t})-\alpha\log\pi_{\theta}\left(\boldsymbol{a}_{t}|\boldsymbol{s}_{t}\right)$.

However, vallina policy gradient methods have several limitations, such as poor data-efficiency and robustness~\citep{schulman2017proximal}. To enhance practical performance, we incorporate this reformulated reward into the PPO framework, leading to the Soft PPO method:
\begin{equation}\label{eqn:Soft PPO obj}
\quad \nabla_\theta J(\pi_\theta) = \nabla_\theta \mathbb{E}_{(\boldsymbol{s}_t,\boldsymbol{a}_t)\sim\pi_\theta,p(\cdot|\boldsymbol{s},\boldsymbol{a})} \left[ \min \{r_t(\pi_\theta) A^{\mathrm{SPG}}_{\pi_\theta}, \operatorname{clip}(r_t(\pi_\theta), 1 - \epsilon, 1 + \epsilon) A^{\mathrm{SPG}}_{\pi_\theta} \} \right].
\end{equation}

% Residual PPO
\textbf{Residual PPO.}
With the derived Soft Policy Gradient in Sec.~\ref{sec:SPG}, the residual policy can be obtained by directly applying it on the augmented MDP $\mathcal{M}^{\mathrm{aug}}=(\mathcal{S}, \mathcal{A},\omega'\log\pi + r_R,p)$:
\begin{equation}
% \label{eqn:Soft PPO obj}
\quad \nabla_\theta J(\pi_\theta) = \nabla_\theta \mathbb{E}_{(\boldsymbol{s}_t,\boldsymbol{a}_t)\sim\pi_\theta,p(\cdot|\boldsymbol{s},\boldsymbol{a})} \left[ \min \{r_t(\pi_\theta) A^{\mathrm{RPG}}_{\pi_\theta}, \operatorname{clip}(r_t(\pi_\theta), 1 - \epsilon, 1 + \epsilon) A^{\mathrm{RPG}}_{\pi_\theta} \} \right].
\end{equation}
where $A^{\mathrm{RPG}}_t$ is estimator of the advantage over the reformulated reward $r^{\mathrm{RPG}}$ at timestep t:

\begin{equation}
\label{eqn: RPG Reward}
r^{\mathrm{RPG}}(\boldsymbol{s}_{t},\boldsymbol{a}_{t}) = 
r_R(\boldsymbol{s}_{t},\boldsymbol{a}_{t}) 
+\omega^\prime \log  {\pi\left(\boldsymbol{a}_{t}|\boldsymbol{s}_{t}\right)} 
- \hat{\alpha} \log {{\pi_{\theta}\left(\boldsymbol{a}_{t}|\boldsymbol{s}_{t}\right)}}.
% = r(\boldsymbol{s}_{t},\boldsymbol{a}_{t}) + \beta \log{\pi\left(\boldsymbol{a}_{t}|\boldsymbol{s}_{t}\right)}- \beta \log {{\pi_{\theta}\left(\boldsymbol{a}_{t}|\boldsymbol{s}_{t}\right)}}
\end{equation}

Building upon the widely used "End-Entropy PPO" (\cite{mnih2016asynchronous}, Eqn.~\ref{eqn: end-entropy}), our method only requires adding the corresponding $- \hat{\alpha} \log {{\pi_{\theta}\left(\boldsymbol{a}_{t}|\boldsymbol{s}_{t}\right)}}$ or $\omega^\prime \log  {\pi\left(\boldsymbol{a}_{t}|\boldsymbol{s}_{t}\right)}$ when computing the advantage, and removes the repeat term $- \alpha \nabla_\theta \mathcal{H}\left(\pi_\theta \left(\cdot | \boldsymbol{s}_t\right) \right)$ when computing the actor loss, to obtain Soft PPO or Residual PPO, making it highly deployable to established PPO pipelines.

% \newpage
\subsection{A Reward View of KL-regularized Objective}

\subsubsection{Not Just KL-regularized Objective}
\label{sec: KL-constrained Reward}
The use of KL objective can be traced back to KL Control~\citep{todorov2006linearly}, which is a branch of stochastic optimal control, and was later introduced into RL~\citep{fox2015taming, rawlik2013stochastic, jaques2017sequence}. The maximized objective follows the form~\citep[Eqn. 2]{jaques2017sequence}:

\begin{equation}
\label{eqn: KL Objective}
\mathbb{E}_{\pi_\theta} \left[ \sum_{t' = t}^{\infty} r_R(\boldsymbol{s}_{t^\prime}, \boldsymbol{a}_{t^\prime}) + \beta \log\frac{ \pi(\boldsymbol{a}_{t^\prime}|\boldsymbol{s}_{t^\prime})}{ \pi_\theta(\boldsymbol{a}_{t^\prime}|\boldsymbol{s}_{t^\prime})} \right],
\end{equation}

% Reward Interpretation
which aligns with the form of maximum-entropy RL objective in Eqn.~\ref{eqn:MERL_obj} by taking $r = r_R + \log \pi$. Follow the derivation in Sec.~\ref{sec:SPG} and Sec.~\ref{sec: practical algorithm}, this form could be view as a special case of RPG whose advantage is caculated on reward:
\begin{equation}
\label{eqn: KL Reward}
r^{\text{KL}}(\boldsymbol{s}_{t},\boldsymbol{a}_{t}) = 
r_R(\boldsymbol{s}_{t},\boldsymbol{a}_{t}) 
+\beta \log  {\pi\left(\boldsymbol{a}_{t}|\boldsymbol{s}_{t}\right)} 
- \beta \log {{\pi_{\theta}\left(\boldsymbol{a}_{t}|\boldsymbol{s}_{t}\right)}}.
% = r(\boldsymbol{s}_{t},\boldsymbol{a}_{t}) + \beta \log{\pi\left(\boldsymbol{a}_{t}|\boldsymbol{s}_{t}\right)}- \beta \log {{\pi_{\theta}\left(\boldsymbol{a}_{t}|\boldsymbol{s}_{t}\right)}}
\end{equation}

Similarly in Eqn.~\ref{eqn: RPG Reward}, the decomposed KL divergence can be viewed as a combination of an augmentation reward $\beta \log \pi$ and an entropy-encouragement term $\beta \log \pi_\theta$. Specifically, if we further assume that prior policy $\pi$ is a maximum-entropy policy encouraged by the same entropy coefficient of the customized policy $\alpha = \beta$, and the optimizing objective is the direct sum of the implicit reward and the preference reward $\omega = 1$, then the KL-regularized objective actually leads to a maximum-entropy policy on a MDP:

\begin{equation}
\label{eqn: KL MDP}
    \mathcal{M}^\mathrm{KL} = (\mathcal{S}, \mathcal{A}, r + r_R, p).
\end{equation}

This decoupling reveals that the KL-constrained reward has a theoretical foundation beyond its simple intuition to just prevent the fine-tuned model from drifting too far from the pretrained model. That is, it is optimizing a trade-off between basic and add-on task on a reward level.

% $r_R(\boldsymbol{s}_{t},\boldsymbol{a}_{t})$ is the add-on reward, $\beta \log \pi$ provided by the prior policy represents the augmented reward, and $\beta \log \pi_\theta$ is the entropy-encouraging term. 

% Moreover, it can be seen as a special case of the proposed RPG reward with $\omega = 1$ and $\alpha = \hat{\alpha} = \beta$.
% explanation
% This implies that Vanilla RLHF is actually optimizing on the equal sum of the implicit reward of human language and a preference reward $r_{\mathrm{human}} + r_R$ ($\omega = 1$), while encouraging the entropy to the same extent ($\alpha = \hat{\alpha}$), or simply:

% Moreover, by assuming that the original LLM was trained under the maximum-entropy principle, this update rule indicates that the RLHF optimizes on a latent full MDP problem $\hat{\mathcal{M}}^\mathrm{RLHF} = (\mathcal{S}, \mathcal{A}, r_{\pi} + r, p)$, where $ r_{\pi}$ is the latent reward the original LLM optimized on.

\subsubsection{Not necessarily a KL-regularized Objective}
\label{sec: More than KL-constrained Reward}
However, according to the derivation of RPG, such reward reparameterization is not limited to being implemented via KL divergence but can instead be decoupled into two tunable parameters $\omega^\prime$ and $\hat{\alpha}$ as in Eqn.~\ref{eqn: RPG Reward}:
$r^{\mathrm{RPG}}(\boldsymbol{s}_{t},\boldsymbol{a}_{t}) = 
r_R(\boldsymbol{s}_{t},\boldsymbol{a}_{t}) 
+\omega^\prime \log  {\pi\left(\boldsymbol{a}_{t}|\boldsymbol{s}_{t}\right)} 
- \hat{\alpha} \log {{\pi_{\theta}\left(\boldsymbol{a}_{t}|\boldsymbol{s}_{t}\right)}}$.

In practice, the prior policy does not necessarily follow the maximum-entropy property, nor is it necessarily optimal. 
Meanwhile, the entropy-encouragement coefficient of the prior policy may not be accessible, and the newly trained policy does not necessarily need to maintain the same.
This flexibility enables fine-tuning without strictly relying on the KL-regularized objective, allowing for a more adaptable approach. 
In the next section, we empirically validate this perspective through experiments by comparing Residual PPO and classic KL-regularized PPO.

In Appendix~\ref{appendix: RLHF}, taking Reinforcement Learning for LLM fine-tuning as an example, we demonstrate how the insights in Sec.~\ref{sec: KL-constrained Reward} and Sec.~\ref{sec: More than KL-constrained Reward} deepen our understanding and informs future directions.

\section{Experiments}

\subsection{Soft PPO}
% \subsubsection{Setting}
\label{sec: Soft PPO}
In this section, we evaluate the preliminary algorithm, Soft PPO, across several continuous control benchmarks in MuJoCo. 

\textbf{Baselines.} We select the variants mentioned in Sec~\ref{sec:SPG} as baselines:
\begin{itemize}[leftmargin=20pt]

\item[-] \textbf{No-Entropy PPO:}
First, we evaluate the classic PPO without entropy term, serving as a baseline performance compared to entropy-regularized approaches.

\item[-] \textbf{End-Entropy PPO:}
Next, we evaluate the form of Eqn.~\ref{eqn: end-entropy} \citep{williams1992simple, mnih2016asynchronous}, which only considers the entropy of the first step. By comparing against this baseline, we investigate whether introducing the entropy term at each step enhances exploration effectiveness.

\item[-] \textbf{Repeat-Entropy PPO:}
Finally, we evaluate the form of Eqn.~\ref{eqn: repeat-entropy} \citep{schulman2017equivalence}, which repeats the entropy of the first step. By comparing against this baseline, we investigate whether removing the repeated entropy term leads to performance improvements.

\end{itemize}

% std design
When modeling the policy distribution, we adopted the common implementation approach in PPO~\citep{mnih2016asynchronous}, using a global standard deviation instead of a state-dependent one. While this reduces the policy interpretability, it improves practical performance by simplifying learning and enhancing stability. A detailed ablation can be found in  appendix~\ref{sec:appendix-Implementation-mujoco}.

% Table Results

\begin{table}[t]
\caption{\textbf{Experimental Results of Entropy-Regularized PPO Variants in MuJoCo.} The evaluation results are over 500 episodes across 5 seeds, and in the form of $\mathrm{mean} \pm \mathrm{std}$.}
\label{tbl: Soft PPO Results}
\begin{threeparttable}
\begin{tabular}{@{}ccccc@{}}

\toprule
\multicolumn{5}{c}{Average Performance} \\
\midrule
\multirow{1}{*}{Env.} & No-Entropy PPO & End-Entropy PPO & Repeat-Entropy PPO & Soft PPO\\ \midrule
Ant          & $\mathbf{3918.9 \pm 1032.9}$ & $3887.9 \pm 991.4$ & $3241.2 \pm 1299.7$ & $3845.6 \pm 911.7$  \\
Hopper       & $2387.3 \pm 1060.6$ & $2696.3 \pm 978.0$ & $2611.6 \pm 967.8$ & $\mathbf{3105.3 \pm 829.5}$  \\
Half Cheetah & $2254.9 \pm 1188.7$ & $3291.5 \pm 2006.0$ & $-114.9 \pm 73.2$ & $\mathbf{4993.7\pm 1262.3}$  \\
\midrule
\multicolumn{5}{c}{Best Performance} \\
\midrule
\multirow{1}{*}{Env.} & No-Entropy PPO & End-Entropy PPO & Repeat-Entropy PPO & Soft PPO\\ \midrule
Ant          & $4389.6 \pm 296.7$ & $4223.4 \pm 822.6$ & $4023.9 \pm 786.3$ & $\mathbf{4581.1 \pm 540.3}$  \\
Hopper       & $3271.2 \pm 573.4$ & $3654.4 \pm 46.4$ & $3404.2 \pm 226.0$ & $\mathbf{3689.4 \pm 29.2}$  \\
Half Cheetah & $4136.6 \pm 1566.6$ & $\mathbf{6049.3 \pm 946.2}$ & $-38.9 \pm 58.2$ & $5896.5\pm 371.8$  \\

\bottomrule

\vspace{-3em} 
\end{tabular}

\end{threeparttable}
\end{table}

% \subsubsection{Results}
\textbf{Does the proposed formulation achieve empirical performance improvements?}
As shown in Table~\ref{tbl: Soft PPO Results}, compared to other variants of entropy-encouraged PPO, the proposed Soft PPO exhibits alike or potentially better average performance. 
Specifically, in the HalfCheetah task, we observed significant performance differences among variants with different entropy encouragement. This indicates that PPO can be very sensitive to the way to encourage entropy-regularized learning: although Repeat-Entropy PPO shares the same parameters and similar updating mechanism with Soft PPO, they can behave completely different during learning.

Notably, the best-performing seeds in the compared variants achieve similar performance to Soft PPO, 
whose overall performance declines are generally due to more seeds converging to the local sub-optimal.
This further indicates that the concise formulation of Soft PPO can potentially better encourage the agent to explore the environment effectively. However, it is also worth noting that since End-Entropy PPO considers entropy encouragement in fundamentally different ways compared to Repeat-Entropy PPO and Soft PPO, they adopt different entropy coefficients, which remains room for better performance with more fine-grained tuning. In practical applications, we consider all these variants to be selectable based on the specific characteristics of the problem.

% \textcolor{entropy-regularized scale issue}

% CPPO Issue
% In PPO, there is a significant issue for practical deployment: even methods that rely on the entropy regularization term often choose, in actual implementation, to set the policy’s standard deviation as a tunable parameter that is independent of the current state and uniform across all actions.

% This assumption leads the log-pi term to reflect less prior policy information, potentially causing performance deterioration in both Soft Policy Learning and Residual Learning. On the other hand, accurately modeling a proper distribution remains a challenging problem in learning, which can likely result in training instability and performance losses under limited computational budget

\subsection{Residual PPO}

% \subsubsection{Setting}

In this section, we evaluate the proposed Residual PPO in the same MuJoCo environments but with extra add-on rewards, which are designed to illustrate customized specifications during practical deployments~\citep{ResidualMPPI}. 
In HalfCheetah, an add-on penalty on the angle of a certain joint is applied to simulate a common issue in deployment when the corresponding motor is broken. 
In Hopper and Ant, the extra rewards are assigned on height and velocity for moving along the y-axis. 
The configurations of environments and training can be found in Appendix~\ref{appendix: environment configuration} and Appendix~\ref{appendix: implementation setting}.

% Metric
\textbf{Metrics.} We evaluate a policy's performance with the total reward $R = r+r_R$, since it represents the objective of the policy customization task. However, when the add-on task conflicts with the basic task, the optimal policy may sacrifice performance on the basic task to maximize the total reward. To better monitor this trade-off during customization, we separately measure the basic and add-on rewards, allowing us to assess how well the customized policy preserves the basic task's performance while optimizing for the add-on task ~\citep{ResidualMPPI}.

% Baselines: Prior/ Residual-MPPI/ Full-MPPI/ Guided-MPPI
\textbf{Baselines.} In our experiments, we mainly compare Residual PPO against four baselines:

\begin{itemize}[leftmargin=20pt]

\item[-] \textbf{Prior Policy:}
First, we evaluate the best-performing Soft PPO checkpoint in Sec.~\ref{sec: Soft PPO} on the full task, which serves as a baseline to show the effectiveness of policy customization.

\item[-] \textbf{Greedy PPO:}
Next, we finetune the prior policy towards the add-on reward solely, which can be achieved by setting $\omega^\prime=0$, which serves as a trival baseline when basic reward is unknown.

\item[-] \textbf{KL PPO:}
Next, we utilize the KL divergence to regularize the fine-tuning process, which can be achieved by setting $\omega^\prime=\hat{\alpha}$, which serves as a widely-used practical fine-tuning baseline~\cite{}.

\item[-] \textbf{Full Policy:}
Finally, we utilize the Soft PPO to train an optimal policy on the full task, which serves as an upper-bound performance of policy customization 

\end{itemize}

% \subsubsection{Results}

\begin{table}[t]
\caption{\textbf{Experimental Results of Residual PPO in MuJoCo.} The evaluation results are over 500 episodes with best-performing checkpoints across 5 seeds. The results are in the form of $\mathrm{mean} \pm \mathrm{std}$.} 
\label{tbl: Customization Results}

\resizebox{\textwidth}{!}{

\begin{threeparttable}
\begin{tabular}{@{}clcccc@{}}
\toprule
\multirow{2}{*}{Env.} & \multicolumn{1}{c}{\multirow{2}{*}{Policy}} & Full Task  & {Basic Task} & \multicolumn{2}{c}{Add-on Task} \\ \cmidrule(l){3-6}
 & & $\mathrm{Total\ Reward}$ & $\mathrm{Basic\ Reward}$ &  $\bar{|\theta|}$ & $\mathrm{Add}$-$\mathrm{on\ Reward}$ \\
 \midrule
 
\multirow{5}{*}{\begin{tabular}[c]{@{}c@{}}Half\\ Cheetah\end{tabular}} 
& Prior Policy  & $5357.0 \pm 377.8$            & $5896.5 \pm 371.8$   & $0.54 \pm 0.01$                &  $-539.8 \pm 8.7$                 \\
\cmidrule(l){2-6} 
& Greedy PPO    & $-1314.0 \pm 252.4$           & $-1150.2 \pm 236.0$  & $0.16 \pm 0.02$                &  $-163.7 \pm 21.8$                \\
& KL PPO        & $5418.7 \pm 150.1$            & $5776.3 \pm 133.7$   & $0.36 \pm 0.02$                &  $-357.6 \pm 17.7$                \\ 
& Residual PPO  & $\mathbf{5488.3 \pm 75.6}$    & $5845.1 \pm 69.5$    & $\mathbf{0.36 \pm 0.01}$       &  $\mathbf{-356.3 \pm 8.6}$        \\
\cmidrule(l){2-6} 
& Full Policy   & $5556.2 \pm 57.8$             & $6011.4 \pm 57.2$    & $0.46 \pm 0.01$                &  $-455.2 \pm 5.8$                 \\
\midrule 
 
\multicolumn{1}{c}{Env} & \multicolumn{1}{c}{Policy} & $\mathrm{Total\ Reward}$ & $\mathrm{Basic\ Reward}$ & $\bar{z}$ & $\mathrm{Add}$-$\mathrm{on\ Reward}$ \\
\midrule
\multirow{5}{*}{Hopper} 
& Prior Policy      & $5133.7 \pm 33.2$             & $3689.4 \pm 29.2$    & $1.44 \pm 0.01$          &  $1444.3 \pm 6.0$             \\
\cmidrule(l){2-6} 
& Greedy PPO        & $3714.5 \pm 7.1$              & $2224.2 \pm 6.5$     & $1.49 \pm 0.00$            &  $1490.2 \pm 0.9$             \\
& KL PPO            & $3864.4 \pm 50.10$            & $2400.3 \pm 55.2$    & $1.46 \pm 0.01$            &  $1464.1\pm 6.1$              \\ 
& Residual PPO      & $\mathbf{4401.4 \pm 505.3}$   & $2951.6 \pm 333.5$   & $\mathbf{1.50 \pm 0.03}$   &  $\mathbf{1449.5 \pm 177.0}$  \\
\cmidrule(l){2-6} 
& Full Policy       & $5142.7 \pm 34.8$             & $3669.0 \pm 26.8$    & $1.47 \pm 0.00$            &  $1473.6 \pm 8.4$             \\
\midrule      
                           
\multicolumn{1}{c}{Env.} & \multicolumn{1}{c}{Policy} & $\mathrm{Total\ Reward}$ & $\mathrm{Basic\ Reward}$ & $\bar{v}_y$ & $\mathrm{Add}$-$\mathrm{on\ Reward}$ \\
\midrule
\multirow{5}{*}{Ant} 
& Prior Policy      & $4875.2 \pm 589.4$            & $4581.1 \pm 540.3$   & $0.29 \pm 0.13$            &  $294.1 \pm 133.9$             \\
\cmidrule(l){2-6} 
& Greedy PPO        & $4340.6 \pm 1464.5$           & $726.4 \pm 277.1$    & $3.86 \pm 0.96$            &  $3613.9 \pm 1246.4$           \\ 
& KL PPO            & $4019.4 \pm 1103.4$           & $368.6 \pm 151.8$    & $3.74 \pm 0.92$            &  $3650.6 \pm 1047.6$           \\ 
& Residual PPO      & $\mathbf{5568.3 \pm 852.4}$   & $1340.8 \pm 220.0$   & $\mathbf{4.27 \pm 0.63}$   &  $\mathbf{4227.4 \pm 705.3}$   \\
\cmidrule(l){2-6} 
& Full Policy       & $6185.1 \pm 1142.0$           & $3543.0 \pm 621.2$   & $2.70 \pm 0.41$            &  $2641.6 \pm 542.8$            \\
\bottomrule

\end{tabular}
    % \begin{tablenotes}
    %     \small
    %     \item 
    % \end{tablenotes}
\end{threeparttable}
}
% \vspace{-2em}
\end{table}

\textbf{Can Residual PPO effectively solve the policy customization problem?}
The experimental results, summarized in Table~\ref{tbl: Customization Results}, demonstrate the effectiveness of Residual PPO for policy customization. 
% Add-on Task Improvements
Across all tasks, Residual PPO achieves obvious performance improvements on the add-on task compared to the prior policy, demonstrating its capability to optimize the prior policy for new requirements. 
Moreover, in the HalfCheetah and Ant, Residual PPO achieves performance comparable or alike performance of the Full Policy, which indicates that by incorporating the $\log \pi$, Residual PPO can effectively balance the trade-off between the basic and the add-on task.

\textbf{Does the flexibility of decoupling lead to more effective policy customization?}
Greedy PPO, due to optimizing towards the add-on reward solely, struggles to maintain its performance on the basic task, resulting in a lower total reward compared to Residual PPO. 
This issue becomes more pronounced in tasks requiring auxiliary rewards, such as action regularization, where Greedy PPO fails to train an effective policy in the HalfCheetah task. 
KL PPO, in theory, should achieve the best performance as it optimizes a MDP with the exact objective $R = r+r_R$ and a reasonable entropy coefficient $\hat{\alpha} = \alpha$. However, as discussed in Sec.~\ref{sec: More than KL-constrained Reward}, the maximum-entropy and optimality assumptions of RPG do not hold in practical scenarios, which can lead to even worse performance, as shown in Ant results.
% Emphasis on Flexibility
In contrast, the decoupling of the KL-regularized term in RPG provides greater tuning flexibility, which can lead to superior performance in practical applications.
% Limitation
However, when this flexibility encounters excessively large assumption misalignments, it can also lead to suboptimal customization results, as observed in the Hopper task.
% Limitation and Future work
This also suggests that when building a foundation model that is more adaptable, adequate modeling of complex distributions is an important direction for future work.

\section{Related Work}\label{sec:related-work}

\textbf{KL-regularized Objective.}
The KL-regularized objective is related to KL Control~\citep{todorov2006linearly, kappen2012optimal}, which aims to prevent the target policy from deviating too far from a reference, which is widely adopted in RL. $\Psi$-learning~\citep{rawlik2013stochastic} and G-learning~\citep{fox2015taming} both maximize the reward while incurring penalties for the KL-divergence from a reference policies.
% RL-Finetuning
The KL-regularized objective is also a common approach in transfer learning~\citep{tirumala2019exploiting} and RL-based finetuning~\citep{nair2020awac}, especially for fine-tuning LLM with human preference. Using RL for LLM fine-tuning as an example, we provide a detailed discussion in Appendix~\ref{appendix: RLHF}.

The flexibility of KL decoupling discussed in Sec.~\ref{sec: More than KL-constrained Reward} can be achieved through alternative approaches. Distral~\citep{teh2017distral} simultaneously employs KL divergence and entropy with different coefficients to regularize the joint training of the task-specific and meta policy. TD-M(PC)\textsuperscript{2}~\citep{lin2025td}
uses the log distribution to bring the nominal policy and planner closer during learning, leading to a more performant policy.
However, these results are derived from practical usage, whereas RPG provides a reward-level perspective to support such implementations.

% \subsection{Residual Q-Learning}
\textbf{Residual Q-Learning.} 
% RQL introduction again
RQL~\citep{RQL} proposes a principled framework to solve policy customization without the reward knowledge of the prior policy. 
% Extension
However, this framework is not limited to RL-based approaches or solely addressing the policy customization: inspired by RQL, Residual-MPPI~\citep{ResidualMPPI} integrates model-based planning to enable online policy customization, while MEReQ~\citep{MEREQ} leverages RQL as an efficient approach for inverse RL.

% Alike Work
Many works originating from preference-based RL have ultimately derived results similar to RQL, and converge toward maximum-entropy RL and specifically the $\log \pi$ form. IQ-Learn~\citep{garg2021iq} and IPL~\citep{hejna2024inverse}  all recognized that learning the maximum-entropy policy or Q-value provides an implicit yet efficient way to capture preferences from data, leading to an updating approach similar to RQL. However, these methods focus on Inverse RL, whereas RQL directly formulates a training pipeline from the perspective of optimizing a new policy.

% \subsection{Model as Reward}
% \textbf{Model as Reward.} 
% % More General Way
% The core of RQL-inspired $\mathcal{M}^\mathrm{aug}$ is utilizing the learned policy as a reward function. This idea has been reflected in many recent works.
% % Introduction
% Beyond the works~\citep{yuan2024implicitprm, rafailov2024direct, rafailov2024r} mentioned in Section~\ref{sec: RQL View}, many other methods focus on a high-level or intuition-driven approach to achieve this goal. 
% % LLM as reward
% \citet{kwon2023reward} first proposed that LLM can work as a reward generator in text-based RL environments. Constitutional AI~\citep{bai2022constitutional} mix preference labels generated by LLMs and humans during the RLHF process.
% % RL-VLM-F: Use VLM as a Labeler
% Furthermore, RL-VLM-F~\citep{wang2024rl} completely replaces human annotators with a VLM to provide preference labels, thus learning a specific reward function.
% % However
% However, despite their practical success, the intuition-driven nature of these methods prevents a deeper insight into what these approaches are actually optimizing.

\section{Conclusion}\label{sec:conclusion}
% Derivation
In this work, we derived a concise form of the maximum-entropy policy gradient and, based on this foundation, developed the \emph{Residual Policy Gradient} as an important complement to RQL. 
% Experiments
With MuJoCo experiments, we validated the theoretical correctness of the proposed algorithms.
% KL-Reward
Moreover, RPG reveals that the KL-constrained reward has theoretical foundation beyond its simple intuition.
% RLHF
This enables us to establish clear connections between many successful RLHF methods and further explain certain experimental phenomena observed in them.
% Future
The success of these works also serves as a strong empirical validation for our theoretical framework, laying a solid foundation for future RLHF and broader policy customization research.

%%%%%%%%%%%%%%%%%%%%%%%%%%%%%%%%%%%%%%%%%%%%%%%%%%%%%%%%%%%%%%%%
%% Appendices
%%%%%%%%%%%%%%%%%%%%%%%%%%%%%%%%%%%%%%%%%%%%%%%%%%%%%%%%%%%%%%%%
\newpage
\appendix

\section{Soft Policy Gradient Derivation}
\label{appendix: derivation}
Let $\tau = (\boldsymbol{s}_0, \boldsymbol{a}_0, \boldsymbol{s}_1, \boldsymbol{a}_1, \dots, \boldsymbol{s}_t, \boldsymbol{a}_t, \dots)$ represents the complete trajectory, and $p_\theta(\tau)$ represents the probability density of $\tau$ generated by a paramterized policy $\pi_\theta$ and the environment dynamics $p(\cdot|\boldsymbol{s},\boldsymbol{a})$. The maximum-entropy objective can be written as:

\begin{equation}\label{eqn: this_MERL_obj}
J(\pi_\theta) = \mathbb{E}_{\tau \sim p_\theta(\tau)}\left[\sum_{t=0} \gamma^t \left(r(\boldsymbol{s}_t,\boldsymbol{a}_t)+\alpha \mathbb{E}_{\boldsymbol{a}_t\sim\pi_\theta(\cdot|\boldsymbol{s}_t)}\left[-\log\pi_\theta\left(\boldsymbol{a}_t|\boldsymbol{s}_t\right)\right]\right)\right].
\end{equation}

Note $\mathbb{E}_{\tau \sim p_\theta(\tau)}\left[\log\pi_\theta\left(\boldsymbol{a}_t|\boldsymbol{s}_t\right)\right] = \mathbb{E}_{\tau \sim p_\theta(\tau)}[\mathbb{E}_{\tau \sim p_\theta(\tau)}\left[\log\pi_\theta\left(\boldsymbol{a}_t|\boldsymbol{s}_t\right)\right]]$
Eqn.~\ref{eqn: this_MERL_obj} can be reformulated as:
\begin{equation}\label{eqn:EqMERL_obj}
J(\pi_\theta) = \mathbb{E}_{\tau \sim p_\theta(\tau)}\left[\sum_{t=0} \gamma^t \left(r(\boldsymbol{s}_t,\boldsymbol{a}_t)  -  \alpha \log\pi_\theta\left(\boldsymbol{a}_t|\boldsymbol{s}_t\right)\right)\right].
\end{equation}

which is equivalent to the standard RL objective with a reward defined as $r(\boldsymbol{s}_t,\boldsymbol{a}_t)-\alpha\log\pi_\theta\left(\boldsymbol{a}_t|\boldsymbol{s}_t\right)$. By defining $R(\tau) =\sum_{t=0} \gamma^t \left(r(\boldsymbol{s}_t,\boldsymbol{a}_t)-\alpha\log\pi_\theta\left(\boldsymbol{a}_t|\boldsymbol{s}_t\right)\right)$, the objective can be rewritten as:
\begin{equation}\label{eqn:PG_obj}
    J(\pi_\theta) = \int p_\theta(\tau) R(\tau) d\boldsymbol{\tau}.
\end{equation}

By taking gradient over $\theta$, we can obtain:
\begin{equation}\label{eqn:PG}
    \nabla_\theta  J(\pi_\theta) = \int \nabla_\theta p_\theta(\tau) R(\tau) + p_\theta(\tau) \nabla_\theta R(\tau) d\boldsymbol{\tau}.
\end{equation}

Due to the Markov properties of the system and policy, the gradient of the trajectory probability density can be written as:
\begin{equation}
\begin{aligned}
        \nabla_\theta p_\theta(\tau) 
        &= p_\theta(\tau) \nabla_\theta \log p_\theta(\tau)\\
        &= p_\theta(\tau) \nabla_\theta \left( \log p(\boldsymbol{s}_0) + \sum_{t=0} \log p(\boldsymbol{s}_{t+1} | \boldsymbol{a}_t, \boldsymbol{s}_t) + \sum_{t=0}\log \pi_\theta(\boldsymbol{a}_t|\boldsymbol{s}_t)\right) \\
        &= p_\theta(\tau)\sum_{t=0} \nabla_\theta \log \pi_\theta(\boldsymbol{a}_t|\boldsymbol{s}_t) \\
\end{aligned}
\end{equation}

Inspired by \cite{thomas2014bias}, we drop $\gamma^t$ for the reduced weight to increase data efficiency and ignore the rewards before time $t$ for each $\nabla_\theta \log \pi_\theta\left(\boldsymbol{a}_t|\boldsymbol{s}_t\right)$ to reduce variance . Therefore, we can simplify the second term:

\begin{equation}\label{eqn: second term}
\begin{aligned}
    &\quad \int p_\theta(\tau) \nabla_\theta R(\tau) d\boldsymbol{\tau} \\
    &= \int p_\theta(\tau) \left(\sum_{t=0} -\alpha \gamma^t \nabla_\theta \log\pi_{\theta}\left(\boldsymbol{a}_t|\boldsymbol{s}_t\right)\right) d\boldsymbol{\tau} \\
    &\approx -\alpha \int p_\theta(\tau) \left(\sum_{t=0} \nabla_\theta \log\pi_{\theta}\left(\boldsymbol{a}_t|\boldsymbol{s}_t\right)\right) d\boldsymbol{\tau} \\
    &= -\alpha \int \nabla_\theta p_\theta(\tau)  d\boldsymbol{\tau} \\
    &= -\alpha \nabla_\theta \int p_\theta(\tau)  d\boldsymbol{\tau} \\
    &= -\alpha \nabla_\theta 1 \\
    &= 0,
\end{aligned}
\end{equation}

and then we apply the same result on Eqn.~\ref{eqn:PG}:
\begin{subequations}\label{eqn:Soft PG}
\begin{align}
    & \nabla_\theta J(\pi_\theta) \\
    = & \int \nabla_\theta p_\theta(\tau) R(\tau) d\boldsymbol{\tau} \\
    = & \int p_\theta(\tau) \left(\sum_{t=0} \nabla_\theta \log \pi_\theta\left(\boldsymbol{a}_t|\boldsymbol{s}_t\right)\right) R(\tau) d\boldsymbol{\tau}\\
    = & \int p_\theta(\tau) \left(\sum_{t=0} \nabla_\theta \log \pi_\theta\left(\boldsymbol{a}_t|\boldsymbol{s}_t\right) \left(\sum_{t^\prime = 0} \gamma^{t^\prime} \left(
    r(\boldsymbol{s}_{t^\prime},\boldsymbol{a}_{t^\prime})-\alpha\log\pi_{\theta}\left(\boldsymbol{a}_{t^\prime}|\boldsymbol{s}_{t^\prime}\right)\right) \right)  \right) d\boldsymbol{\tau} \\
    \approx & \int p_\theta(\tau) \left(\sum_{t=0} \nabla_\theta \log \pi_\theta\left(\boldsymbol{a}_t|\boldsymbol{s}_t\right) \left(\sum_{t^\prime = t} \gamma^{t^\prime - t} \left(
    r(\boldsymbol{s}_{t^\prime},\boldsymbol{a}_{t^\prime})-\alpha\log\pi_{\theta}\left(\boldsymbol{a}_{t^\prime}|\boldsymbol{s}_{t^\prime}\right)\right) \right)  \right) d\boldsymbol{\tau} \\
    = & \mathbb{E}_{\tau \sim p_\theta(\tau)} \left[ \sum_{t=0} \nabla_\theta \log \pi_\theta\left(\boldsymbol{a}_t|\boldsymbol{s}_t\right) \left(\sum_{t^\prime = t} \gamma^{t^\prime - t} \left(
    r(\boldsymbol{s}_{t^\prime},\boldsymbol{a}_{t^\prime})-\alpha\log\pi_{\theta}\left(\boldsymbol{a}_{t^\prime}|\boldsymbol{s}_{t^\prime}\right)\right) \right) \right]
\end{align}
\end{subequations}

\section{Example: Reinforcement Learning for LLM Fine-tuning}
\label{appendix: RLHF}
% Introduce RLHF
Reinforcement Learning for LLM fine-tuning represents a canonical instance of policy customization, where the exact reward model of the original LLM is unknown. However, the objective remains to fine-tune the LLM to align with new preferences while preserving its prior conversational capabilities. Here we show how the insights in Sec.~\ref{sec: KL-constrained Reward} and Sec.~\ref{sec: More than KL-constrained Reward} deepen our understanding and informs future research.

Vanilla RLHF~\citep[Eqn. 7]{ziegler2019fine} follows the intuition of~\cite{jaques2017sequence}, adding a KL regularization term to reward when computing advantage, as:
\begin{equation}
\label{eqn: RLHF Reward}
r^{\mathrm{RLHF}}(\boldsymbol{s}_{t},\boldsymbol{a}_{t}) = 
r_R(\boldsymbol{s}_{t},\boldsymbol{a}_{t}) 
+\beta \log  {\pi\left(\boldsymbol{a}_{t}|\boldsymbol{s}_{t}\right)} 
- \beta \log {{\pi_{\theta}\left(\boldsymbol{a}_{t}|\boldsymbol{s}_{t}\right)}}.
\end{equation}

As we analyzed in previous sections, if we assume that the distribution of human language can be represented as a Boltzmann distribution from an implicit reward $r_{\mathrm{human}}$, and that the original LLM models an approximately optimal distribution, then Vanilla RLHF is actually optimizing on a MDP:
\begin{equation}
\label{eqn: RLHF MDP}
    \mathcal{M}^\mathrm{RLHF} = (\mathcal{S}, \mathcal{A}, r_{\mathrm{human}} + r_R, p).
\end{equation}

Direct Preference Optimization (DPO)~\citep{rafailov2024direct, rafailov2024r} also realizes the importance of reward-level modeling. It defines a reparameterized reward as:
\begin{equation}
\label{eqn: DPO Reward}
     r^{\mathrm{DPO}}\left(\boldsymbol{s}_{t}, \boldsymbol{a}_{t}\right) = \beta \log \frac{\pi_\theta\left(\boldsymbol{a}_{t}|\boldsymbol{s}_{t}\right)}{\pi\left(\boldsymbol{a}_{t}|\boldsymbol{s}_{t}\right)},
\end{equation}
and a reward loss of preference demonstrations $\{x^{(i)}, y^{(i)}_w, y^{(i)}_l\}_{i=1}^N$ under Bradley-Terry model:
\begin{equation}
\label{eqn: DPO Loss}
\mathcal{L}_{\mathrm{DPO}}(\pi_{\theta}; \pi) =  \log \sigma \left( \beta \log \frac{\pi_{\theta}(y_w | x)}{\pi(y_w | x)} - \beta \log \frac{\pi_{\theta}(y_l | x)}{\pi(y_l | x)} \right) ,
\end{equation}

thus transforming policy learning into reward learning from demonstrations. From the view of RPG, the reparameterization idea of DPO could be easily derived by modeling the log full policy $\beta \log \pi_\theta({\boldsymbol{s}_{t}| \boldsymbol{a}_{t}})$ as the total reward function $R \left(\boldsymbol{s}_{t}, \boldsymbol{a}_{t}\right)$:

% Explain why cross-entropy loss
\begin{equation}
\label{eqn: DPO from RPG}
r_R({\boldsymbol{s}_{t}, \boldsymbol{a}_{t}}) = 
R \left(\boldsymbol{s}_{t}, \boldsymbol{a}_{t}\right) - r_{\mathrm{human}} \left(\boldsymbol{s}_{t}, \boldsymbol{a}_{t}\right) =
 \beta \log \frac{\pi_\theta\left(\boldsymbol{a}_{t}|\boldsymbol{s}_{t}\right)}{\pi\left(\boldsymbol{a}_{t}|\boldsymbol{s}_{t}\right)} =  r^{\mathrm{DPO}}\left(\boldsymbol{s}_{t}, \boldsymbol{a}_{t}\right).
\end{equation}

Moreover, with the insights of RPG, we can identify several potential impovervements starting from DPO. First, modeling $\log \pi$ as reward requires it to be a maximum-entropy policy, which means that a proper reward loss that aligns with this assumption, e.g., cross-entropy loss, can lead to potentially better performance:
\begin{equation}
\label{eqn: CrossEntropy-DPO}
\mathcal{L}_{\mathrm{CrossEntropy-DPO}} = l \cdot \log \sigma \left( \beta \log \frac{\pi_{\theta}(y_w | x)}{\pi(y_w | x)} \right) 
+ (1 - l) \cdot \log \left[ 1 - \sigma \left( \beta \log \frac{\pi_{\theta}(y_l | x)}{\pi(y_l | x)} \right) \right].
\end{equation}
% The experimental results observed in a recent work, Implicit PRM~\citep{yuan2024implicitprm}, exactly support our insights by showing that, when modeling $\pi_\theta$, using cross-entropy as the loss function demonstrates particularly higher data efficiency compared to other loss functions.

The experimental results from a recent work, Implicit PRM~\citep{yuan2024implicitprm}, provide supporting evidence for our insights: when modeling $\pi_\theta$, using cross-entropy as the loss function tends to yield higher data efficiency compared to other loss functions.

Second, the reparameterized reward does not have to be a KL-regularized objective in practice, which means that we can develop a decomposed form of DPO loss with more flexibility:
\begin{equation}
\label{eqn: Decomposed-DPO}
\mathcal{L}_{\mathrm{Decomposed-DPO}}(\pi_{\theta}; \pi) = \log \sigma \left( 
\hat{\alpha} \log \frac{\pi_{\theta}(y_w | x)}{\pi_{\theta}(y_l | x)}
- \omega^\prime \log \frac{\pi(y_w | x)}{\pi(y_l | x)} \right).
\end{equation}
Recently, Q-Adapter~\citep{li2024q}, an RLHF framework based on RQL, has provided further validation of this approach by demonstrating that adjusting the value of $\omega^\prime$ can effectively balance the trade-off and even enhance performance on both the basic and add-on tasks in RLHF.
Furthermore, Q-Adapter strictly follows the RQL formulation, integrating the $\log \pi$ term into the value function update rather than incorporating it in an augmented manner at each step. This design improves data efficiency and enhances training stability, a key aspect also discussed in RQL~\citep{RQL}.

\section{MuJoCo Experiments}

\subsection{Environment Configuration} 
\label{appendix: environment configuration}
In this section, we introduce the detailed configurations of the selected environments, including the basic tasks, add-on tasks, and the corresponding rewards. The action and observation space of all the environments follow the default settings in \texttt{gym[mujoco]-v3}.
% All the environments terminate automatically after 1000 environmental steps. The state spaces are all organized in the form of positional values of different body parts of the agent, followed by the velocities of those individual parts (their derivatives). 

\textbf{Half Cheetah.} In the HalfCheetah environment, the basic goal is to apply torque on the joints to make the cheetah run forward (right) as fast as possible. 
The state and action space has 17 and 6 dimensions, and the action represents the torques applied between links.

The basic reward consists of two parts: forward reward $r_\mathrm{forward}(\boldsymbol{s}_t, \boldsymbol{a}_t) = \frac{\Delta x}{\Delta t}$ and control cost $r_\mathrm{control}(\boldsymbol{s}_t, \boldsymbol{a}_t) = -0.1 \times \
||\boldsymbol{a}_t||_2^2$.
During policy customization, we demand an additional task that requires the cheetah to limit the angle of its hind leg. 
This customization goal is formulated as an add-on reward function defined as $r_R(\boldsymbol{s}_t, \boldsymbol{a}_t) = -|\theta_{\mathrm{hind \ leg}}|$

\textbf{Hopper.} 
In the Hopper environment, the basic goal is to make the hopper move in the forward direction by applying torques on the three hinges connecting the four body parts.
The state and action space has 11 and 3 dimensions, and the action represents the torques applied between links.

The basic reward consists of three parts: alive reward $r_\mathrm{alive} = 1$, forward reward $r_\mathrm{forward}(\boldsymbol{s}_t, \boldsymbol{a}_t) = \frac{\Delta x}{\Delta t}$, control cost $r_\mathrm{control}(\boldsymbol{s}_t, \boldsymbol{a}_t) = -0.001 \times ||\boldsymbol{a}_t||_2^2$.
During policy customization, we demand an additional task that requires the hopper to jump higher along the z-axis.
This customization goal is formulated as an add-on reward function defined as $r_R(\boldsymbol{s}_t, \boldsymbol{a}_t) = z $.

\textbf{Ant.} 
In the Ant environment, the basic goal is to coordinate the four legs to move in the forward (right) direction by applying torques on the eight hinges connecting the two links of each leg and the torso (nine parts and eight hinges).
The state and action space has 27 and 8 dimensions, and the action represents the torques applied at the hinge joints.

The basic reward consists of three parts: alive reward $r_\mathrm{alive} = 1$, forward reward $r_\mathrm{forward}(\boldsymbol{s}_t, \boldsymbol{a}_t) = \frac{\Delta x}{\Delta t}$, control cost $r_\mathrm{control}(\boldsymbol{s}_t, \boldsymbol{a}_t) = -0.5 \times ||\boldsymbol{a}_t||_2^2$.
During policy customization, we demand an additional task that requires the ant to move along the y-axis. This customization goal is formulated as an add-on reward function defined as $r_R(\boldsymbol{s}_t, \boldsymbol{a}_t) = \frac{\Delta y}{\Delta t}$.

\subsection{Implementation Setting}
\label{appendix: implementation setting}
The training pipeline is developed upon rsl-rl~\citep{rudin2022learning} implementation. All variants are prior policies and customized policies share the task-specific parameters finetuned from the predefined settings in this implementation
% , which can be found in the provided code. 
The policy customization process is executed for 5M steps in the environment, where the actor model is directly loaded from the prior policy. Additionally, the first 5\% of steps are used to fit a reinitialized value function to stabilize the training process.
The primary differences among variants lie in the choice of entropy coefficient and augmentation coefficient, which can be found in Table~\ref{tbl: Entropy-Encourage Coefficient} and Table~\ref{tbl: Augmentation Coefficient}. 
% All experimental results in the paper can be reproduced by setting the seeds in the provided code.

\begin{table*}[h]
      \centering
      \caption{Entropy-Encourage Coefficient $\alpha$ in Prior Policy Training}
      \scalebox{1}{
      \begin{tabular}{cccccc}
        \toprule

         $\alpha$ & Half Cheetah & Ant  & Hopper &  \\
        \midrule
        End-Entropy PPO         & $0.01$     & $0.001$   & $0.0005$  \\
        Repeat-Entropy PPO      & $0.13472$  & $0.0001$  & $0.001$   \\
        Soft PPO                & $0.13472$  & $0.0001$  & $0.001$   \\

        \bottomrule
      \end{tabular}}
      \label{tbl: Entropy-Encourage Coefficient}
\end{table*}

\begin{table*}[h]
      \centering
      \caption{Augmentation Coefficient $\omega^\prime$ in Policy Customization}
      \scalebox{1}{
      \begin{tabular}{cccccc}
        \toprule

         $\omega^\prime$ & Half Cheetah & Ant  & Hopper &  \\
        \midrule
        Greedy PPO         & $0$     & $0$   & $0$  \\
        KL PPO             & $0.13472$  & $0.0001$  & $0.001$   \\
        Residual PPO       & $0.17$     & $0.01$   & $0.01$ \\

        \bottomrule
      \end{tabular}}
      \label{tbl: Augmentation Coefficient}
\end{table*}

\subsection{Global Std. Ablation}

\begin{threeparttable}
\caption{\textbf{Ablation of policy distribution modeling in MuJoCo Experiments.} The evaluation results are computed over 500 episodes across 5 seeds. The results are in the form of $\mathrm{mean} \pm \mathrm{std}$ }
\begin{tabular}{@{}ccccc@{}}

\toprule
\multirow{2}{*}{Env.} & \multicolumn{2}{c}{Average Performance}  & \multicolumn{2}{c}{Best Performance}\\ 

\cmidrule(l){2-5}
             & Global Std. &  State Std. & Global Std. & State Std.\\
\midrule
Ant          & $\mathbf{3845.6 \pm 911.7}$ & $3055.3 \pm 1231.5$ & $\mathbf{4581.1 \pm 540.3}$ & $3915.1 \pm 599.9$  \\
Hopper       & $\mathbf{3105.3 \pm 829.5}$ & $1504.0 \pm 948.3$ & $\mathbf{3689.4 \pm 29.2}$ & $2511.6 \pm 784.6$  \\
Half Cheetah & $\mathbf{4993.7\pm 1262.3}$  & $2621.7 \pm 1605.3$ & $\mathbf{5896.5 \pm 371.8}$ & $5405.0\pm 1288.8$  \\

\bottomrule
\label{tbl: std ablation}
\end{tabular}
\end{threeparttable}

% std design
As shown in Table~\ref{tbl: std ablation}, despite the sacrifices in policy interpretability, the widely adopted global std design outperforms the state-dependent std variant in both average performance and best performance in practical applications.

\label{sec:appendix-Implementation-mujoco}

% \section{Extra Experiments on Legged Gym}
% \section*{Appendix}
% % No label, since this can't be referenced meaningfully with \ref{}.
% This format should only be used if there is a single appendix (unlike in this document).

%%%%%%%%%%%%%%%%%%%%%%%%%%%%%%%%%%%%%%%%%%%%%%%%%%%%%%%%%%%%%%%%
%% NOTE: THIS MARKS THE END OF THE "MAIN TEXT"
%%%%%%%%%%%%%%%%%%%%%%%%%%%%%%%%%%%%%%%%%%%%%%%%%%%%%%%%%%%%%%%%

%%%%%%%%%%%%%%%%%%%%%%%%%%%%%%%%%%%%%%%%%%%%%%%%%%%%%%%%%%%%%%%%
%% Bibliography
%%%%%%%%%%%%%%%%%%%%%%%%%%%%%%%%%%%%%%%%%%%%%%%%%%%%%%%%%%%%%%%%
\newpage
\bibliography{main}
\bibliographystyle{rlj}

%%%%%%%%%%%%%%%%%%%%%%%%%%%%%%%%%%%%%%%%%%%%%%%%%%%%%%%%%%%%%%%%
% AUTHOR: If your paper has no supplementary materials, you may 
%         comment out the line below, which creates the title for
%         the supplementary materials.
%%%%%%%%%%%%%%%%%%%%%%%%%%%%%%%%%%%%%%%%%%%%%%%%%%%%%%%%%%%%%%%%
% \beginSupplementaryMaterials

% Content that appears after the references are not part of the ``main text,'' have no page limits, are not necessarily reviewed, and should not contain any claims or material central to the paper. 
% %
% If your paper includes supplementary materials, use the \begin{center}
%     {\tt {\textbackslash}beginSupplementaryMaterials} 
% \end{center}
% command as in this example, which produces the title and disclaimer above. 
% %
% If your paper does not include supplementary materials, this command can be removed or commented out.

\end{document}